\documentclass[letterpaper, 10 pt, conference]{ieeeconf}  % Comment this line out if you need a4paper

\usepackage{graphicx}
\usepackage{subcaption}
\usepackage{booktabs}
\usepackage{pifont}
\usepackage{titlesec}

\IEEEoverridecommandlockouts                              % This command is only needed if 
\title{\LARGE \bf
Behavior Cloning for Active Perception with Low-Resolution Egocentric Vision
}

\author{Anthony Bilic$^1$, Chen Chen$^1$, and Ladislau Bölöni$^1$
\thanks{$^1$Department of Computer Science, University of Central Florida, Orlando, FL, USA}}
\begin{document}

\maketitle
\thispagestyle{empty}
\pagestyle{empty}

%%%%%%%%%%%%%%%%%%%%%%%%%%%%%%%%%%%%%%%%%%%%%%%%%%%%%%%%%%%%%%%%%%%%%%%%%%%%%%%%
\begin{abstract}
We investigate whether behavior cloning is sufficient to produce active perception in a structured object-finding task. A low-cost robot arm equipped with a wrist-mounted egocentric RGB camera must reposition to center a partially visible plant before triggering a grasp signal, requiring actions that improve future observations. The model predicts joint commands directly from low-resolution RGB images under closed-loop control. We show that low-resolution egocentric vision is sufficient for reliable task completion and that predicting relative joint deltas substantially outperforms absolute joint position prediction in our setting. These results demonstrate that visually grounded active perception can emerge from behavior cloning in a reproducible setting.
\end{abstract}

%%%%%%%%%%%%%%%%%%%%%%%%%%%%%%%%%%%%%%%%%%%%%%%%%%%%%%%%%%%%%%%%%%%%%%%%%%%%%%%%
\section{Introduction}
This work investigates whether behavior cloning is sufficient to produce active perception in a structured object-finding task. Behavior cloning learns policies by imitating expert demonstrations~\cite{hafner2019learning, osa2018algorithmic, chi2024universal, rahmatizadeh2018vision} and does not explicitly optimize for information-seeking actions. Active perception, by contrast, concerns systems in which actions are selected to deliberately influence future observations in order to enable task completion~\cite{ballard1991animate, bajcsy2018revisiting}.

To evaluate this question, a controlled object-finding experiment is constructed using a low-cost robot arm equipped with a wrist-mounted egocentric RGB camera. A target plant is initially only partially visible, requiring the robot to reposition the camera until the object is centered and fully observable before triggering a grasp signal. Successful completion therefore requires actions that improve subsequent visual input.

A neural network is trained to map low-resolution RGB observations to joint commands under closed-loop control. We consider two representation for the joint commands: absolute joint positions~\cite{ rahmatizadeh2018vision, zhao2023learning} and joint deltas~\cite{chi2024universal, levine2018learning, zhu2022viola}, where a joint delta denotes the change from the current joint configuration to the next. The impact of this action representation choice is evaluated empirically for our task.

We make the following contributions:
\begin{itemize}
\item We propose a simple and reproducible experimental setup using a low-cost robot arm equipped with a wrist-mounted egocentric RGB camera to evaluate active perception.
\item We demonstrate that behavior cloning produces active perception behavior without explicit supervision for information gathering.
\item We show that low-resolution egocentric RGB input is sufficient for reliable active perception task completion under closed-loop control.
\item We show that predicting relative joint deltas yields substantially better performance, smoothness, and generalization than absolute joint position prediction in this setting.
\end{itemize}

\section{Method}

\subsection{Task Definition and Setup}

\begin{figure}[t]
    \centering
    
    \begin{subfigure}[t]{0.48\linewidth}
        \centering
        \includegraphics[height=4cm]{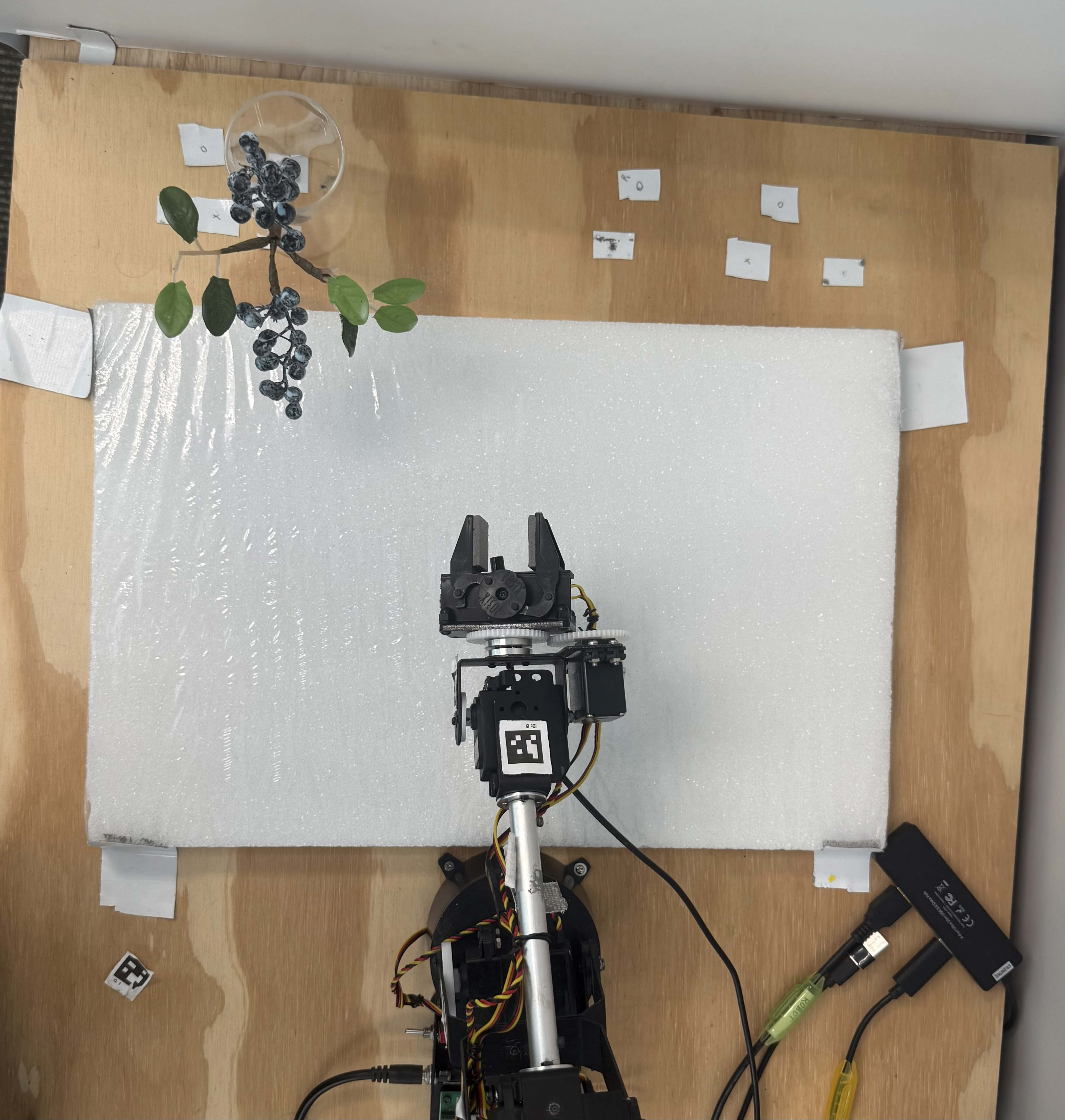}
    \end{subfigure}
    \hfill
    \begin{subfigure}[t]{0.48\linewidth}
        \centering
        \includegraphics[height=4cm]{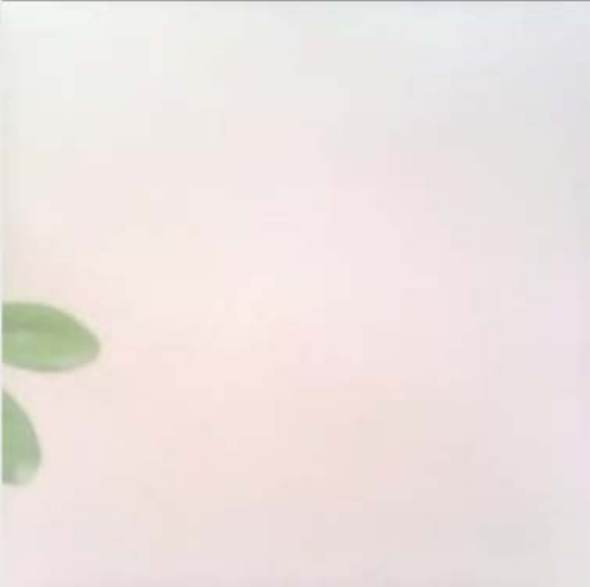}
    \end{subfigure}
    
    \caption{
    Left: Overhead view of experimental setup. Right: Initial observation of wrist-mounted view with the plant on the left.
    }
    \vspace{-5mm}
    \label{fig:setup_simplified}
\end{figure}

Experiments are conducted using a table-mounted Lynxmotion AL5D robotic arm with 6 degrees of freedom and a wrist-mounted monocular RGB camera (Fig.~\ref{fig:setup_simplified}). The system is built from inexpensive off-the-shelf components. Demonstrations are collected via teleoperation using an Xbox controller. The wrist camera captures 64$\times$64 RGB images at $10$ Hz, and joint positions are recorded synchronously at the same rate.

At the start of each demonstration, a plant is placed to the left or right such that only part of it is visible against an uninformative white background (Fig.~\ref{fig:setup_simplified}). The task is to center the plant in view and close the gripper. Because the plant is not fully visible at initialization, the robot must execute movements that change the wrist camera viewpoint until the object becomes centered. Actions therefore influence subsequent observations, providing a controlled evaluation of active perception in an intentionally minimal and reproducible setting.

\begin{figure*}[!t]
    \centering
    \includegraphics[width=\textwidth]{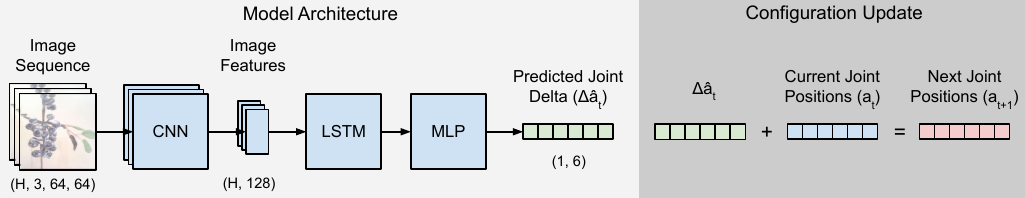}
    \caption{Model architecture (left) and joint configuration update (right).}
    \label{fig:architecture}
    \vspace{-5mm}
\end{figure*}

\subsection{Approach}

We train a visual encoder and temporal controller end-to-end using behavior cloning (Fig.~\ref{fig:architecture}). The visual encoder is a four-layer convolutional neural network (CNN) that processes individual RGB images to produce compact feature representations, leveraging spatial inductive bias for efficient perception. The controller is an LSTM that processes these features over time, integrating information across timesteps to capture dependencies that are not observable from a single frame.

We use a fixed-length history of images as input to the model. Training data is constructed by sliding a window of $H$ frames across teleoperated demonstrations, where each window defines a history ending at timestep $t$. The supervision target for each history is the corresponding demonstrated joint delta $\Delta a_t = a_{t+1} - a_t$, representing the change in joint positions from the final observation in the history to the next timestep. The model is trained to predict joint deltas by minimizing mean squared error (MSE) between predicted and demonstrated joint deltas, $\mathrm{MSE}(\hat{\Delta a}_t, \Delta a_t)$.

During inference, the history is initialized by repeating the first observed image $H$ times. Once initialized, the model predicts a joint delta from the current history. This delta is added to the robot’s current joint positions to obtain $a_{t+1} = a_t + \Delta a_t$, which the robot executes to update its configuration. A new image is then captured from the updated state and appended to the history while the oldest is removed, maintaining a fixed-length input of $H$ images, and the process repeats. To mirror this process during training, each demonstration is padded at the start by repeating the initial frame $H - 1$ times when constructing input histories.

%%%%%%%%%%%%%%%%%%%%%%%%%%%%%%%%%%%%%%%%%%%%%%%%%%%%%%%%%%%%%%%%%%%%%%%%%%%%%%%%
\section{Experiments}

Table~\ref{tab:delta_vs_position} reports the mean squared error (MSE) of joint predictions, computed in joint position space, for both joint delta and absolute position prediction across different numbers of demonstrations, together with task success rates. In the absolute position setting, the model is trained identically but with targets defined as $a_{t+1}$ instead of $a_{t+1} - a_t$, as is commonly done in prior work. Demonstrations are divided into an 80--10--10 train/validation/test split, where training set sizes are sampled from the training portion while the validation and test sets remain fixed across experiments. Demonstrations consist of trials where the plant is placed on the left or right with equal probability, and splits are balanced accordingly. At test time, the model runs autonomously in closed loop for 10 seconds with $H = 20$. A trial is considered successful if the robot stabilizes on the plant and closes the gripper exactly once. Early or repeated closures are counted as failures.

With eight demonstrations, both models complete the task. With four demonstrations, only the delta model succeeds. Training with joint deltas yields significantly lower test loss, which is reflected during execution. The delta model makes smaller and more consistent movements toward the target, whereas the absolute position models make larger movements and frequently overshoot before correcting. Both model's behavior becomes smoother as the number of demonstrations increases.

We also test plant placements between the fixed left and right training positions. The delta model adapts to these unseen placements when the plant is partially visible at initialization. The absolute position model instead moves toward one of the demonstrated left or right configurations once the plant comes into view. This difference stems from the action representation. The position model learns to move to a specific joint configuration given an image, whereas the delta model learns to produce a relative adjustment from its current state. This explains why delta prediction better supports adaptation to intermediate placements.

\begin{table}[t]
\centering
\caption{Mean squared error (MSE) of joint predictions (in joint position space) for joint delta and absolute position prediction across different training set sizes (Demo Count). MSE values are scaled by $10^{-3}$.}
\label{tab:delta_vs_position}
\begin{tabular}{c rr c rr c}
\toprule
& \multicolumn{3}{c}{Delta} & \multicolumn{3}{c}{Position} \\
\cmidrule(lr){2-4} \cmidrule(lr){5-7}
Demo Count & Train & Test & Success & Train & Test & Success \\
\midrule
2  & 5.16 & 6.22 & 0/5 & 17.13 & 157.48 & 0/5 \\
4  & 5.23 & 6.15 & 5/5 & 13.23 & 189.08 & 0/5 \\
8  & 5.69 & 6.08 & 5/5 & 9.73  & 93.28  & 5/5 \\
16 & 5.65 & 5.91 & 5/5 & 12.84 & 30.52  & 5/5 \\
32 & 5.08 & 5.60 & 5/5 & 9.20  & 16.57  & 5/5 \\
64 & 5.36 & 5.54 & 5/5 & 8.71  & 12.80  & 5/5 \\
\bottomrule
\end{tabular}
\vspace{-4mm}
\end{table}
%%%%%%%%%%%%%%%%%%%%%%%%%%%%%%%%%%%%%%%%%%%%%%%%%%%%%%%%%%%%%%%%%%%%%%%%%%%%%%%%
\section{Conclusion}

We present a simple and reproducible setup for evaluating whether behavior cloning is sufficient for active perception. In this object-finding task, the learned model produces closed-loop behavior that repositions the camera to improve future observations. Low-resolution egocentric RGB input is sufficient for reliable task completion, and predicting relative joint deltas yields more stable and generalizable behavior than absolute joint position prediction. These results show that visually grounded active perception can emerge from behavior cloning in this setting.

\noindent\textbf{Acknowledgment:} This work was partly supported by the intramural research program of the U.S. Department of Agriculture, National Institute of Food and Agriculture via grant number 2024-67022-41788

 \clearpage


%%%%%%%%%%%%%%%%%%%%%%%%%%%%%%%%%%%%%%%%%%%%%%%%%%%%%%%%%%%%%%%%%%%%%%%%%%%%%%%%

%%%%%%%%%%%%%%%%%%%%%%%%%%%%%%%%%%%%%%%%%%%%%%%%%%%%%%%%%%%%%%%%%%%%%%%%%%%%%%%%



% \section*{ACKNOWLEDGMENT}

% The preferred spelling of the word ÒacknowledgmentÓ in America is without an ÒeÓ after the ÒgÓ. Avoid the stilted expression, ÒOne of us (R. B. G.) thanks . . .Ó  Instead, try ÒR. B. G. thanksÓ. Put sponsor acknowledgments in the unnumbered footnote on the first page.

%%%%%%%%%%%%%%%%%%%%%%%%%%%%%%%%%%%%%%%%%%%%%%%%%%%%%%%%%%%%%%%%%%%%%%%%%%%%%%%%

% References are important to the reader; therefore, each citation must be complete and correct. If at all possible, references should be commonly available publications.

% \clearpage

\bibliographystyle{IEEEtran}
\bibliography{root}

\end{document}